# Real-Time Detection and Tracking of Foreign Object Intrusions in Power Systems via Feature-Based Edge Intelligence

Xinan Wang, *Member, IEEE*, Di Shi, *Senior Member, IEEE*, Fengyu Wang, *Senior Member, IEEE*

*Abstract*— This paper presents a novel three-stage framework for real-time foreign object intrusion (FOI) detection and tracking in power transmission systems. The framework integrates: (i) a YOLOv7 segmentation model for fast and robust object localization, (ii) a ConvNeXt-based feature extractor trained with triplet loss to generate discriminative embeddings, and (iii) a feature-assisted IoU tracker that ensures resilient multi-object tracking under occlusion and motion. To enable scalable field deployment, the pipeline is optimized for deployment on low-cost edge hardware using mixed-precision inference. The system supports incremental updates by adding embeddings from previously unseen objects into a reference database without requiring model retraining. Extensive experiments on real-world surveillance and drone video datasets demonstrate the framework's high accuracy and robustness across diverse FOI scenarios. In addition, hardware benchmarks on NVIDIA Jetson devices confirm the framework's practicality and scalability for real-world edge applications.

## I. Introduction

Foreign object intrusion (FOI) poses a serious and growing threat to the safe and stable operation of power transmission systems. FOIs can cause equipment failure, service interruptions, and even catastrophic accidents. Despite advances in monitoring technologies, reliably detecting FOIs—particularly in real time—remains a critical challenge.

FOI events can be broadly categorized into three types: naturally induced, intentional, and accidental. Each presents unique detection challenges and requires different monitoring strategies.

Naturally induced FOIs are caused by environmental factors such as vegetation encroachment, bird nests, or fallen branches. These are typically static and can be detected using periodic inspections with surveillance cameras or sensor networks. Prior research has demonstrated promising results using deep learning methods, e.g., CNN- and YOLO-based models for nest detection [1], [2], Faster R-CNN for vegetation monitoring [3], and LiDAR-based 3D reconstructions for encroachment analysis [4], [5]. Other recent work has applied noise-separation networks for static FOI detection [6], [7].

Intentional FOIs, including theft, vandalism, or sabotage, often require integrated solutions that combine surveillance, anomaly detection, and security policies. For instance, [7], [8] proposed video analytics and IoT-based detection systems, while [9], [10] explored microgrid reconfiguration to enhance resilience.

Accidental FOIs—such as construction equipment, wind-blown debris, or agricultural materials—fundamentally differ from the other two categories. They occur without warning, escalate rapidly, and can lead to fires, outages, or equipment damage [11], [12]. Unlike naturally induced FOIs that evolve gradually over hours or days, accidental FOIs demand continuous, real-time monitoring and immediate response. Several challenges hinder reliable detection in this context: 1) unpredictable human activity near transmission lines makes object motion difficult to model; 2) rapid onset leaves little time for intervention; many existing anomaly detection methods are designed for offline use; and 3) high intra-class variance—including deformable or irregular debris—limits the generalization of conventional detectors [13], [14].

Some studies have attempted to combine detection with tracking. For example, YOLO-based models have been adapted for construction vehicle monitoring [16], CNNs with oriented bounding boxes for motion tracking [17], and time-series features for abnormal activity detection [18], [19]. However, these methods fall short of addressing the combined challenges of unpredictability, deformability, and deployment under edge constraints.

This paper therefore focuses on the challenge of real-time accidental FOI detection and tracking in dynamic, unstructured environments. To overcome these limitations, we propose a robust three-stage framework optimized for edge deployment in power transmission systems. The system integrates 1) a YOLOv7-based segmentation model for rapid object localization, 2) a ConvNeXt-based feature encoder trained with triplet loss for generalizable embeddings, and 3) a feature-assisted IoU-based tracker for reliable multi-object association under motion and occlusion.

The novelty of this work lies in the following aspects: First, unlike prior single-stage YOLO/Faster R-CNN approaches, our framework decouples segmentation from classification and employs feature-based retrieval, enabling retraining-free scalability when new FOI categories emerge. Second, the integration of ConvNeXt embeddings with triplet loss provides robust discrimination under severe intra-class variance, a scenario poorly handled by conventional detectors. Third, the entire pipeline is explicitly designed for real-time operation on low-cost edge hardware (e.g., Jetson Orin Nano), achieving practical scalability for large-scale deployment in the field.

Extensive experiments using surveillance and drone footage validate the effectiveness of the proposed method under diverse and challenging conditions, and hardware benchmarking demonstrates its suitability for edge deployment.

The rest of this paper is organized as follows: Section II describes the detection and tracking algorithm, Section III presents the edge deployment strategy, Section IV provides case studies with real-world data, and Section V concludes with future directions.

## II. THE PROPOSED ALGORITHM

Accidental FOIs from sources such as drones, construction equipment, and wind-blown debris present significant challenges for detection, especially when involving non-rigid objects like dust-proof nets, greenhouse films, etc. Although YOLO variants (e.g., YOLOv5, YOLOv8) are efficient, their performance often degrades significantly on low-power edge devices due to their high memory and compute requirements when deployed with lower precision weights [21]. Faster R-CNN, with its two-stage architecture, is even more computationally intensive and generally unsuitable for real-time inference on edge hardware like the Jetson Orin Nano [22]. Both YOLO and Faster R-CNN rely on fixed classifier heads trained on a limited object vocabulary. They struggle to generalize when deployed in open-set conditions or when encountering unseen objects, especially deformable or wind-blown ones [23]. This makes them less suitable for dynamic FOI scenarios, which involve a wide range of shapes, motions, and occlusion levels. In addition, adding new object types in traditional YOLO/Faster R-CNN pipelines requires expensive retraining and fine-tuning. This is impractical in real-world transmission line monitoring where new FOI types (e.g., different wind-blown banners or construction materials) can appear frequently.

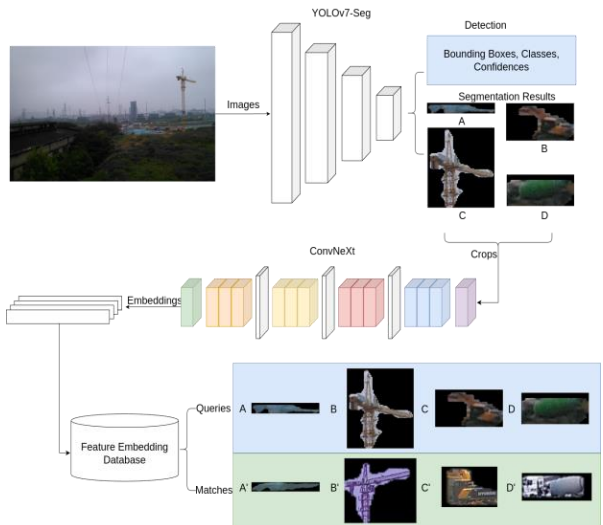

Fig. 1 The proposed three-stage foreign object detection method

To overcome these limitations, we propose a three-stage detection and tracking framework as shown in Fig. 1. The first stage uses a YOLOv7 segmentation model [24] to extract object masks from complex backgrounds. For example, in Fig. 1, four foreign objects, dust-proof net, tower crane, excavator, and cement mixer, are identified and cropped (A–D). The second stage applies a fine-tuned ConvNeXt model to extract 1×1024 feature embeddings that encode discriminative visual information. In the third stage, each embedding is compared to a pre-constructed reference dataset using cosine similarity, and the closest match determines the object class. This feature similarity-based matching mechanism allows rapid updates by adding embeddings, without retraining the model. Fig. 1 shows how the query crops (A–D) are assigned to their corresponding reference embeddings (A'–D'). The subsequent sections detail the components of this framework.

### A. YOLOv7 Segmentation with Class Aggregation

This study focuses on several common FOI types including greenhouse films, dust-proof nets, wind-blown banners, metal roof sheets, tower cranes, crane vehicles, cement mixers, excavators, and cement pumps. Image data were collected from surveillance cameras mounted on transmission towers. Due to variations in camera location, angle, and height, the images exhibit significant scale diversity and strong class imbalance.

YOLO-based models are optimized for fast object localization and real-time detection. However, when trained on datasets with high intra-class variance and uneven class distributions, their classification performance degrades significantly [21], [23]. In such settings, a single-stage detection model struggles to generalize across visually inconsistent object types. To overcome this limitation, we design a three-stage framework in which the YOLOv7 segmentation model [24] serves only as the object localizer, while classification is delegated to a feature extractor in the second stage.

We choose segmentation over detection for two reasons: 1) *improved noise isolation*: segmentation masks help isolate foreground objects from background noise, improving downstream feature extraction; 2) *robustness to object deformation*: non-rigid objects, which often exhibit shape distortion and variable boundary contours, are better localized using mask-based segmentation. We summarized the performance of the most popular segmentation models in Table 1. It can be seen that among all the candidates, YOLOv7-seg achieves the best balance between accuracy and inference speed compared to Mask R-CNN [25], RTMDet [26], DeepLabV3+, SAM, YOLACT and Fast-SCNN. This makes it well-suited for edge deployment.

TABLE 1 BENCHMARK COMPARISON BETWEEN POPULAR SEGMENTATION MODELS

| Model | Architecture | Speed (FPS) | Accuracy (mAP) | Edge | Multi-task (Det+Seg) |
|---|---|---|---|---|---|
| **YOLOv7-seg** | One-stage | 20–40 (Orin Nano) | High | Excellent | Yes |
| **RTMDet-Seg** | Anchor-free, one-stage | 20–40 (Orin Nano) | High | Good | Yes |
| **Mask R-CNN** | Two-stage | ~5–10 (GPU) | Very High | Bad | Yes |
| **DeepLabV3+** | Encoder-Decoder | ~20–30 (GPU) | High | Moderate | No |
| SAM | Transformer-based | ~1–2 (GPU) | Very High | Bad | No |
| YOLACT | One-stage | ~30–50 (GPU) | Moderate | Good | Yes |
| Fast-SCNN | Encoder Decoder | 60+ (Mobile GPU) | Low | Excellent | No |

To further enhance localization, we adopt a class aggregation strategy, as supported in [27], which simplifies the model's classification task and allows more capacity to be devoted to bounding box and mask optimization. In our case, three different class aggregation schemas are proposed based on the 10 classes' visual similarities, and the one with the highest mean average precision (mAP) in training is used. The three schemas are:

- Grouping by material: a) metal objects: all construction machines and metal roof sheet; b) mesh objects: dust-proof net; c) plastic objects: wind-blown banner and greenhouse film.
- Grouping by height: a) high objects: Crane tower, cement pump, crane vehicle; b) medium height objects: Excavator, cement mixer, bulldozer, metal roof sheet; c) ground-contact objects: Dust-proof net, greenhouse film, wind-blown banner.
- Grouping by functional behavior: a) non-rigid objects: greenhouse films, dust-proof nets, wind-blown banners; b) construction machinery: tower cranes, crane vehicles, cement mixers, excavators, bulldozers, cement pumps; c) rigid objects: metal roof sheets.

TABLE 2 MEAN AVERAGE PRECISION ACROSS DIFFERENT CLASS AGGREGATIONS

| Aggregation Strategy | mAP@0.5 | Precision | Recall |
|---|---|---|---|
| Grouping by material | 72.4% | 74.1% | 70.8% |
| Grouping by height | 75.5% | 77.0% | 74.3% |
| Grouping by functional behavior | 78.2% | 80.3% | 76.0% |

Table 2 shows the ablation experiment results, and the functional behavior-based class aggregation shows the best performance in terms of mAP@0.5, precision and recall. So we decided to aggregate the original 10 classes into 3 broader categories: 1) non-rigid objects: greenhouse films, dust-proof nets, wind-blown banners; 2) construction machinery: tower cranes, crane vehicles, cement mixers, excavators, bulldozers, cement pumps; 3) rigid objects: metal roof sheets.

This aggregation reduces the total number of classes $N$ and improves learning stability for rare or highly variable object types. To understand the impact of class reduction, we analyze the YOLOv7-seg loss and its gradient behavior. The total training loss is:

$$L_{total} = L_{class} + L_{box} + \alpha L_{seg\_bce} + \beta L_{seg\_dice} \quad (1)$$

Here, $L_{class}$ is the multi-label classification loss, $L_{box}$ measures bounding box accuracy, $L_{seg\_bce}$ is the pixel-wise binary cross-entropy loss for foreground-background segmentation, $L_{seg\_dice}$ ensures accurate shape alignment, and $\alpha$ and $\beta$ are loss weighting coefficients. The total gradient with respect to the model weights $W$ is:

$$\frac{\partial L_{total}}{\partial W} = \frac{\partial L_{class}}{\partial W} + \frac{\partial L_{box}}{\partial W} + \alpha \frac{\partial L_{seg\_bce}}{\partial W} + \beta \frac{\partial L_{seg\_dice}}{\partial W} \quad (2)$$

The classification loss $L_{class}$ uses cross-entropy (BCE) for multi-label classification:

$$L_{class} = -\frac{1}{N}\sum_{i=1}^{N}[y_i log(p_i) + (1-y_i)log(1-p_i)] \quad (3)$$

Applying the chain rule;

$$\frac{\partial L_{class}}{\partial W} = \sum_{i=1}^{N} \frac{\partial L_{class}}{\partial p_i} \cdot \frac{\partial p_i}{\partial z_i} \cdot \frac{\partial z_i}{\partial W} \quad (4)$$

Assuming $p_i=\sigma(z_i)$ and $z_i=w^T x+b$, we derive:

$$\frac{\partial L_{class}}{\partial W} = \frac{1}{N}\sum_{i=1}^{N}(p_i - y_i) \cdot x \quad (5)$$

This shows that the number of trainable parameters influenced by $L_{class}$ grows with $N$, the number of classes. Reducing $N$ reduces the complexity of the classification space, which in turn shifts optimization capacity toward improving mask boundaries and object localization. In summary, class aggregation in the segmentation stage improves training efficiency, reduces overfitting, and strengthens model performance under class imbalance—particularly for diverse and non-rigid FOI categories.

### B. Foreign Object Feature Embedding Extraction

After obtaining object mask crops from the YOLOv7 segmentation model, a feature extraction network is used to encode visual characteristics of each object into high-dimensional embeddings. These embeddings enable comparison and classification in later stages without relying on conventional classification heads.

Many state-of-the-art feature extraction models originate from image classification architectures trained on large datasets such as ImageNet. These models, including ResNet, EfficientNet, ConvNeXt, and Vision Transformers (ViTs), learn rich visual representations that can be repurposed for other tasks by removing the final classification layer.

Among the most widely used architectures are Vision Transformers (ViT) [28], EfficientNet-L2 [29], and ConvNeXt [18]. ViT offers high accuracy on large datasets but is computationally expensive and slow during inference. EfficientNet is lightweight and well-suited for edge devices with limited data and computing resources. ConvNeXt, a modernized convolutional neural network that incorporates design elements from transformers (e.g., large kernels and LayerNorm) demonstrated superior feature extraction and classification capabilities compared to traditional backbones used in detection models (e.g., [18], [30], [31]). This enhanced representational power enables better discrimination of subtle intra-class differences and robustness under challenging conditions. In addition, it offers a balance between speed, accuracy, and efficiency. Table summarizes the key characteristics of these three models in terms of computational needs and performance.

TABLE 3 COMPARISON OF FEATURE EXTRACTION ARCHITECTURE

| Model | Inference Speed | Training Cost | Computing Cost | Data Requirement | Accuracy |
|---|---|---|---|---|---|
| **EfficientNet-L2** | Fast | Moderate | Low | Small/Medium | Moderate |
| **ViT** | Slow | Very High | High | Massive | High |
| **ConvNeXt** | Fast | Moderate | Moderate | Small/Medium | Good |

Given our requirement for real-time video inference on edge devices, ConvNeXt is selected as the feature extractor due to its strong performance on medium-sized datasets, competitive accuracy, and fast inference speed.

By default, ConvNeXt is trained using the cross-entropy loss for multi-class classification. However, cross-entropy does not explicitly structure the feature space to ensure separation among different object classes. To generate embeddings that are both discriminative and generalizable, we adopt the triplet loss, which encourages similar objects to cluster together while pushing apart embeddings from different classes. The triplet loss function operates on three inputs: anchor $a$, positive sample $p$ from the same class, and negative sample $n$ from a different class. The loss is defined as:

$$L = max(0, \|f(a) - f(p)\|_2^2 - \|f(a) - f(n)\|_2^2 + margin) \quad (6)$$

Here, $f(\cdot)$ denotes the embedding output, and margin is a small positive constant. This encourages embeddings of the same class to be closer than those of different classes by at least the specified margin. The loss reaches zero when:

$$\|f(a) - f(p)\|_2^2 + margin < \|f(a) - f(n)\|_2^2 \quad (7)$$

In this work, we use 1024-dimensional embeddings and cosine similarity as the metric for comparing vectors during inference.

To adapt ConvNeXt for embedding learning, the final classification layer is replaced with a fully connected layer that outputs 1024-dimensional embeddings. This updated architecture is illustrated in Fig. 2.

During training, the dataset is organized into triplets $[a, p, n]$, where $a$ and $p$ belong to the same class and $n$ belongs to a different one. To address data imbalance, triplets are sampled evenly across all classes. Each training iteration involves three forward passes (one per sample in the triplet), computation of the triplet loss via (6), and one backpropagation step to update the model weights. This training strategy enables the model to learn a structured embedding space that supports plug-and-play inference across a variety of foreign object types.

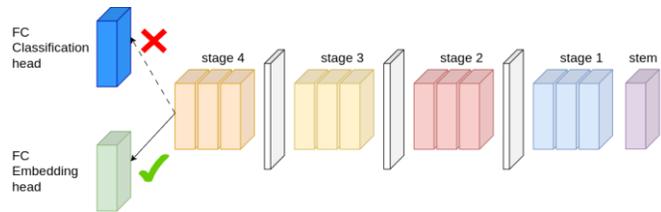

Fig. 2 Modified ConvNeXt model for feature embedding extraction

Once training is complete, the ConvNeXt model is used to generate reference embeddings for a labeled image set. These embeddings are stored in a searchable dataset and serve as a fixed comparison base during inference. When a new object is detected, its embedding is computed and compared to this dataset using cosine similarity. The closest match determines its class.

ConvNeXt embeddings, due to their structured convolutional hierarchies and implicit invariances, demonstrate strong separability and robustness for samples from unseen classes or novel conditions. This is evidenced by a clustering-based generalization index on unseen samples [32] and its sensitivity test against data natural distribution shift [33]. These findings demonstrate the system's capability to generalize to unseen object types by adding reference embeddings with new labels without retraining the model, thereby ensuring scalability, ease of maintenance, and robust generalization across deployment sites.

### C. Feature Embedding Dataset Construction

Constructing a feature embedding dataset is essential for enabling classification via embedding comparison in modern computer vision systems. Image embeddings are dense, low-dimensional representations that encode key semantic features such as shape, texture, and structural patterns. These embeddings allow fast and effective comparisons using standard distance metrics, enabling scalable and efficient classification at inference time.

A well-designed embedding dataset offers several key advantages: 1) *efficient similarity search*: embeddings can be compared using cosine similarity or Euclidean distance, allowing real-time inference with minimal computational overhead, 2) *easy extensibility*: new object types can be added by generating and appending their embeddings, without requiring model retraining; 3) *traceability and interpretability*: each embedding is associated with its source image, which enables validation, fine-tuning, or correction of misclassified samples.

All images used to generate the embeddings must be labeled consistently with the training set used to fine-tune the feature extraction model. Each data point in the embedding dataset contains four components: a unique index, a 1024-dimensional feature embedding vector, the file path of the source image, and the class label. This structure is illustrated in Fig. 3.

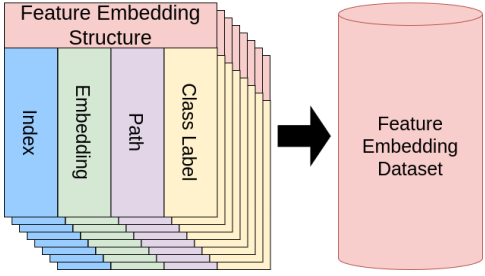

Fig. 3 Data structure of the feature embedding dataset

The choice of search algorithm significantly impacts both classification speed and accuracy. For small or moderate-sized datasets, brute-force search and *k*-nearest neighbors (kNN) [34], [35] offer exact and interpretable results but scale poorly with dataset size. For large-scale datasets, approximate nearest neighbors (ANN) [36] or clustering-based indexing [37] can greatly improve search speed, albeit with minor accuracy trade-offs.

In our implementation, the dataset size remains moderate. We adopt a brute-force search combined with cosine similarity to balance accuracy and inference speed. Cosine similarity between two embedding vectors X and Y is defined as:

$$cosine(X,Y) = \frac{X \cdot Y}{\|X\|\|Y\|} \quad (8)$$

where $X \cdot Y$ is the dot product and $\|X\|$ and $\|Y\|$ are the Euclidean norms.

During tracking, the same object may appear in multiple consecutive frames. To improve classification robustness, we use majority voting based on per-frame embedding matches. Let $\{E_0, E_1, \ldots, E_n\}$ denote the embeddings of the tracked object across $n$ frames, and let $D=\{R_j\}$ be the reference embedding dataset. For each frame $i$, the class label $y_i$ is assigned by finding the closest match in the reference set:

$$y_i = argd(E_i, R_j), R_j \in D \quad (9)$$

where $d(\,,\,)$ is cosine similarity.

The final predicted class for the object is determined by the majority vote:

$$\hat{y} = mode(\{y_0, y_1, \ldots, y_n\}) \quad (10)$$

This approach improves robustness to transient noise, occlusions, or brief tracking errors, and ensures consistent object classification over time.

### D. Pros & Cons Discussion

The proposed YOLOv7-seg + ConvNeXt framework decouples the segmentation/detection stage from the object identification stage. The segmentation network produces generic object masks without being restricted to a predefined set of object categories. Identification is performed by comparing extracted features with entries in a reference feature database.

For unseen FOIs, retraining the detector is not required if the object is visually distinguishable in the segmentation stage; only its representative features need to be added to the reference database.

The system's robustness stems from the use of ConvNeXt features, which capture high-level texture and shape descriptors that generalize beyond the training categories.

But this method also carries practical limits. The scalability without retraining depends on two main factors: (1) segmentation network's ability to produce sufficiently accurate masks for unfamiliar shapes or textures, and (2) the separability of the new object's feature vector from existing entries in the feature space. When the new FOI is visually similar to existing categories or poorly segmented, the feature similarity threshold may lead to misclassification, in which case retraining or fine-tuning may be necessary. However, this retraining is required far less frequently and is significantly less burdensome compared to traditional methods that require retraining of the entire detection backbone whenever new classes are introduced.

In addition, our approach assumes relatively stable camera mounting (fixed or slowly varying viewing angle) and environmental conditions (sufficient lighting, limited severe weather or occlusion) so that segmentation produces reliable object masks. Under extreme glare, heavy rain/snow, or rapid camera motion, the segmentation quality—and therefore downstream identification—may degrade. The recommended edge computing hardware should have sustained throughput of 20 TOPS (INT8) or beyond.

## III. EDGE DEPLOYMENT

### A. Hardware Selection

The proposed framework is designed for real-time operation on edge devices, requiring simultaneous execution of four core components: the YOLOv7 segmentation model, the ConvNeXt feature extraction model, the embedding dataset server, and the object tracking algorithm. To meet the performance and integration requirements of such a pipeline, we selected the NVIDIA Jetson Orin Nano platform based on the following criteria:

1) *Cost-effectiveness*: the Jetson Orin Nano is economically viable, especially for scalable, distributed deployment;

2) *Computational performance*: With up to 67 TOPS, it provides sufficient processing power to support low-latency inference for real-time applications;

3) *Multi-framework compatibility*: through the NVIDIA Triton Inference Server, the platform supports multiple machine learning frameworks and model formats natively, eliminating the need for cross-format conversion;

4) *Multi-model and ensemble support*: the device supports concurrent execution of multiple models and enables ensemble inference pipelines, facilitating modular and efficient deployment.

### B. Model Training and Deployment

A total of 877 images were collected from surveillance cameras installed at over 120 transmission tower sites, each containing instances of FOIs. For training the YOLOv7 segmentation model, the dataset was split into training, validation, and testing sets in a 7:2:1 ratio, yielding 614 training images, 175 validation images, and 88 test images.

To enhance model robustness, basic data augmentation techniques, including horizontal flipping and rotation, were applied, expanding the training set to 1,842 images. Fig. 4 illustrates examples of annotated data used to train the YOLOv7 segmentation model.

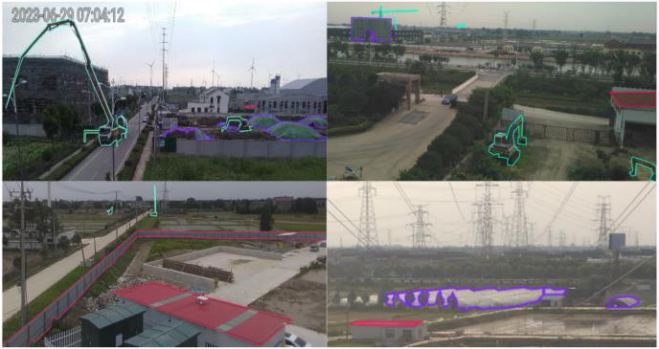

Fig. 4 Object annotation samples in the training images

To evaluate the effect of class aggregation on model performance, an A/B test was conducted using the same dataset. In Scenario A, object labels were grouped into three aggregated classes as described in Section II-A: rigid objects, non-rigid objects, and construction machinery. Scenario B used the original 10-class labeling without aggregation. Fig. 5 compares the training and validation losses in both scenarios. Class aggregation (Scenario A) significantly improved model performance:

- Box loss: decreased from 0.041 to 0.031 (training) and from 0.081 to 0.064 (validation).
- Segmentation loss: reduced from 0.030 to 0.020 (training) and from 0.064 to 0.047 (validation).

These results confirm that class aggregation enhances YOLOv7's localization precision and segmentation quality by simplifying the classification task and reducing overfitting.

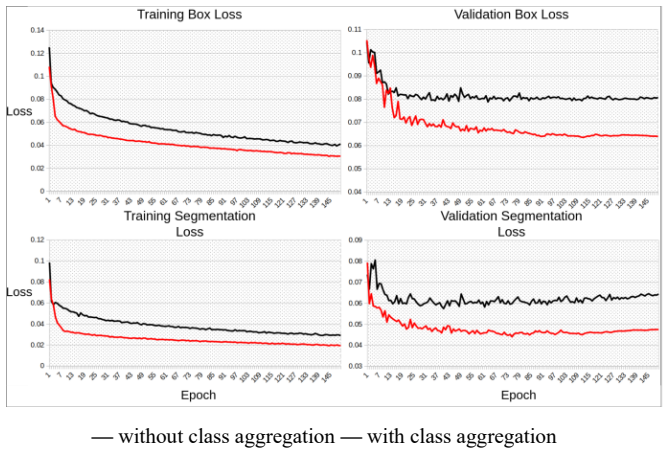

— without class aggregation — with class aggregation

Fig. 5 YOLOv7-seg model training performance comparison

After training the segmentation model, all 877 original images were reprocessed to extract object mask crops for training the ConvNeXt-based feature embedding model. Fig. 6 shows representative samples of cropped masks. The resulting dataset includes: 130 bulldozers, 86 cement mixers, 74 cement pumps, 309 crane vehicles, 339 excavators, 532 crane towers, 1,212 dust-proof nets, 89 wind-blown banners, 788 greenhouse films, and 954 metal roof sheets. These instances were split into training and testing subsets using a 7:3 ratio. From the training subset, 10,000 triplets were generated for training the ConvNeXt model using the triplet loss function described in Section II-B.

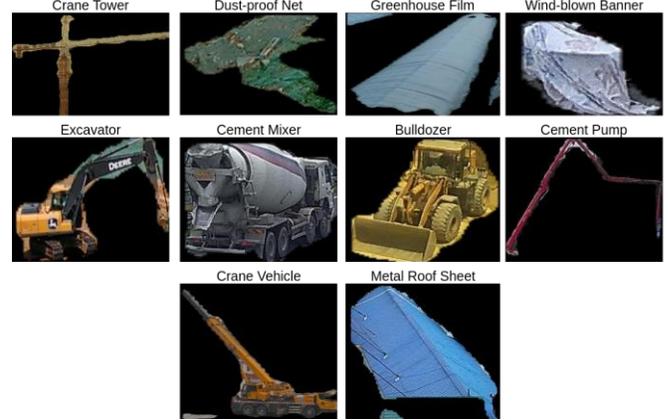

Fig. 6 Object crop mask samples

To evaluate the effectiveness of the proposed YOLOv7-seg (with class aggregation) and ConvNeXt three-stage classification, the testing results were compared with the baseline YOLOv7-seg model without class aggregation. Comparison results are shown in Table 4. Performance metrics include precision, recall, mAP@0.5, and F1 score across all object classes.

$$Precision = \frac{True\ Positive\ (TP)}{True\ Positive\ (TP) + False\ Positive\ (FP)} \quad (11)$$

$$Recall = \frac{True\ Positive\ (TP)}{True\ Positive\ (TP) + False\ Negative\ (FN)} \quad (12)$$

$$F1 = 2 * \frac{Precision * Recall}{Precision + Recall} \quad (13)$$

TABLE 4 COMPARISON OF FEATURE EXTRACTION ARCHITECTURE

| Class | YOLOv7-seg (class aggregated) + ConvNeXt | | | | YOLOv7-seg (without class aggregation) | | | |
|---|---|---|---|---|---|---|---|---|
| Metric | P | R | @0.5 | F1 | P | R | @0.5 | F1 |
| All | 0.8 | 0.82 | 0.83 | 0.85 | 0.58 | 0.47 | 0.52 | 0.51 |
| Bulldozer | 0.94 | 0.82 | | 0.88 | 0.91 | 0.73 | 0.83 | 0.81 |
| Cement mixer | 0.98 | 0.93 | | 0.95 | 0.98 | 0.9 | 0.92 | 0.94 |
| Cement pump | 0.88 | 1.00 | 0.90 | 0.94 | 0.41 | 1.00 | 0.63 | 0.58 |
| Crane vehicle | 1.00 | 0.84 | | 0.91 | 1.00 | 0.82 | 0.85 | 0.90 |
| Excavator | 0.99 | 0.96 | | 0.97 | 1.00 | 0.72 | 0.83 | 0.84 |
| Tower crane | 0.89 | 0.90 | | 0.89 | 0.64 | 0.62 | 0.58 | 0.63 |
| WB Banner | 0.78 | 0.67 | | 0.72 | 0.43 | 0.28 | 0.33 | 0.34 |
| Greenhouse film | 0.81 | 0.78 | 0.79 | 0.79 | 0.45 | 0.34 | 0.41 | 0.39 |
| DP net | 0.91 | 0.76 | | 0.83 | 0.51 | 0.44 | 0.52 | 0.47 |
| Metal roof | 0.83 | 0.83 | 0.81 | 0.83 | 0.40 | 0.22 | 0.29 | 0.28 |

The testing results show that the model YOLOv7-seg with class aggregation + ConvNeXt significantly outperforms YOLOv7-seg without class aggregation across nearly all classes and evaluation metrics, and here are some key findings:
- Overall Performance (All Classes):
  - Precision (P) improves from 0.58 → 0.80
  - Recall (R) improves from 0.47 → 0.82
  - mAP@0.5 improves from 0.52 → 0.83
  - F1 score improves from 0.51 → 0.85
- Heavy Machinery Classes (e.g., Bulldozer, Excavator, Cement Mixer):
  - High precision and recall across both models, but the class aggregation + ConvNeXt model shows more consistent balance, especially in F1 scores (often ≥ 0.90).
- Challenging Object Classes (e.g., Wind-blown Banner, Greenhouse film, Dust-proof net, and Metal roof):
  - The baseline model struggles, with F1 scores as low as 0.28–0.47, while the enhanced model maintains F1 scores in the 0.72–0.83 range.

## C. Embedding Storage and Retrieval

After training the YOLOv7 segmentation model and the ConvNeXt feature extraction model, the latter was applied to generate embeddings for all 4,513 cropped object images, including both training and testing sets. These 1024-dimensional embeddings form the reference feature dataset described in Section II-C and illustrated in Figure 3.

To support real-time retrieval, the entire embedding dataset is hosted on a Redis in-memory database, which provides fast storage and lookup capabilities. Given the moderate size of the dataset, we employ a cosine similarity-based brute-force search strategy for classification. For each query embedding, the system performs an exhaustive comparison against all reference entries and identifies the closest match based on cosine similarity.

This approach eliminates the need for approximate indexing, maintains classification accuracy, and allows flexible dataset updates. Its efficiency and precision make it particularly well-suited for deployment on edge platforms with moderate computational capacity.

## D. Feature-Assisted Multi-Object Tracking

To enable real-time response to FOI events, the proposed framework integrates a robust multi-object tracking mechanism that maintains object identities across video frames. While traditional tracking algorithms rely on Intersection over Union (IoU) to associate detections with existing tracks, IoU-based methods often struggle when objects have low overlap (e.g., fast movement) or are partially occluded, leading to ambiguous or incorrect associations.

To overcome these challenges, we augment the IoU tracker with *appearance-based feature embeddings* extracted by the ConvNeXt model. This hybrid tracking approach combines spatial and visual information to improve object matching in difficult scenarios. The IoU between two bounding boxes $a$ and $b$ is defined as:

$$IOU(a,b) = \frac{Area(a) \cap Area(b)}{Area(a) \cup Area(b)} \quad (13)$$

The proposed feature-assisted IoU tracking process follows the steps below:

*1) Object detection per frame*: The current video frame is processed by the YOLOv7-seg model to detect all objects and output their bounding boxes and segmentation masks.

*2) Initial IoU-based matching:* Detected objects are matched to existing tracks by computing the IoU between each detection and previously tracked objects. The detection is associated with the track that yields the highest IoU (if IoU ≥ 0.5).

*3) Feature embedding extraction:* For each detected object, the cropped mask is passed through the ConvNeXt model to generate a 1024-dimensional feature embedding.

*4) Feature-based disambiguation:* If IoU matching is ambiguous (e.g., multiple overlapping candidates or low IoU scores), the system compares the feature embedding of the detection to candidate tracks using cosine similarity. The track with the highest similarity score is chosen.

*5) Track update and management:* The selected track is updated with the new position and visual features. If no match meets the threshold, a new track is initialized. Tracks that remain unmatched for a set number of frames are marked as inactive or lost.

*6) FOI classification and reporting*: For active tracks approaching sensitive infrastructure, the associated feature embedding is compared with the Redis-hosted reference dataset. The object type is determined via cosine similarity, and if the object poses a risk, an alert is generated for system operators.

## E. Edge Deployment Architecture

Fig. 8 illustrates the deployment architecture of the proposed FOI detection and tracking system on an edge device. The YOLOv7-seg and ConvNeXt models are hosted on an NVIDIA Triton Inference Server and accessed via gRPC [38]. In real-time operation, incoming video frames are first processed by YOLOv7-seg to detect objects and generate bounding boxes along with segmentation masks. The cropped object regions are then forwarded to the ConvNeXt model to extract 1024-dimensional feature embeddings. These embeddings, along with their corresponding bounding boxes, are passed to the tracker, which updates object states and evaluates their motion. If an object appears to be moving toward critical infrastructure, its embedding is compared against the Redis-hosted reference dataset using cosine similarity to determine its class. If a threat is confirmed, an alert is generated and sent to system operators for timely intervention.

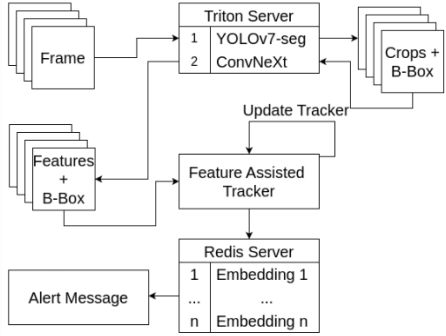

Fig. 8 Logic flow of the proposed three-stage FOI tracking algorithm on edge

This deployment strategy ensures low-latency, fully on-device processing by combining Triton Server for inference, gRPC for inter-component communication, and Redis for high-speed embedding retrieval.

### F. Comparative Analysis with Recent Work

To highlight the distinctions and improvements introduced by our three-stage framework, we present a comparative summary in Table 5. This table outlines key features, performance metrics, and deployment characteristics of the most recent FOI detection systems in other research alongside our method.

These methods include the recent advances that have shown promise in identifying foreign objects in power transmission environments. For instance, Wang et al. [39] improved YOLOv8m with a Global Attention Module to enhance detection accuracy on high-resolution aerial imagery, while in [40], authors proposed a YOLOv8 Network with Bidirectional Feature Pyramid Network (YOLOv8_BiFPN), which can effectively detect foreign objects on power transmission lines from different scales through passing a broader level of feature maps to the whole network. Sun et al. [41] integrate the Swin Transformer (ST) into the YOLOv8 backbone to improve the model's feature extraction capabilities for foreign object detection. The results show that the proposed YOLOv7 segmentation + ConvNeXt method demonstrates strong real-time performance on edge hardware (Jetson Orin Nano) while remaining extendable to new object classes. Although its mAP@0.5 (88.5%) is slightly lower than other models, it is based on tasks with more classes (10) and is the only approach that offers deployment on edge devices. In contrast, prior methods such as YOLOv8m+GAM and YOLOv8+BiFPN achieved higher accuracy (95.5% and 90.2%, respectively) but lack class extensibility and rely on high-end GPUs (e.g., RTX 2080 Ti, 4060 Ti), making them less suitable for real-world edge applications.

TABLE 5 COMPARISON OF FEATURE EXTRACTION ARCHITECTURE

| Backb-one | Extendable | Class No | mAP@0.5 | FPS | Hardware | Dataset | Edge |
|---|---|---|---|---|---|---|---|
| YOLOv8m +GAM [39] | — | 6 | 95.5% | 27.9 | RTX 2080 Ti | 3.8k | — |
| YOLOv8 +BiFPN [40] | — | 6 | 90.2% | n/a | n/a | 1.5k | — |
| YOLOv8 +ST [41] | — | 3 | 89.7% | 181.4 | RTX 4060 Ti | 1.1k | — |
| YOLOv7seg +ConvNeXt | ✓ | 10 | 88.5% | ≈20 | Orin Nano | <1k | ✓ |

## IV. CASE STUDY

To demonstrate the computational efficiency of the proposed FOI detection and tracking framework, we tested our framework using two popular NVIDIA edge devices: Jetson Orin Nano 8GB and Jetson AGX Orin 32GB. The key metrics such as inference speed (FPS), CPU/GPU utilization, memory footprint, and power consumption under typical workloads are compared and presented in Table 6.

TABLE 6 END-TO-END HARDWARE BENCHMARK COMPARISON

| Metric | Orin Nano | AGX Orin |
|---|---|---|
| YOLOv7 Seg Inference | 26–45 ms/frame | 12–22 ms/frame |
| ConvNeXt (2–3 crops) | 15–25 ms/frame | 6–10 ms/frame |
| Cosine + Redis lookup | ~1 ms/feature | <1 ms/feature |
| Total Latency per Frame | 42–73 ms | 18–35 ms |
| Estimated FPS (end-to-end) | 14–24 FPS | 29–56 FPS |
| Memory Usage | ~2.2 GB | ~3.5–4.5 GB |
| GPU Utilization | ~70–85% | ~50–65% |
| CPU Utilization | ~15–25% (4-core) | ~5–10% (12-core) |
| Power Consumption | ~9–15 W | ~18–45 W |

The proposed framework comprising a YOLOv7 segmentation model (TensorRT FP16), a ConvNeXt feature extractor (ONNX FP32), and Redis-based cosine similarity search—achieved near real-time inference speed on Jetson Orin Nano and real-time inference speed on Jetson AGX Orin. As shown in Table 3, with almost 4 times higher unit price, the AGX Orin offers substantial headroom in both compute and memory, making it suitable for high-throughput or multi-camera applications, while the Orin Nano delivers sufficient performance for lightweight, cost- and power-constrained edge deployments. These results highlight the trade-off between computational performance and unit cost. In this research, most foreign object movements are relatively slow; therefore, we choose the Jetson Orin Nano 8GB to form a low-cost solution for the FOI detection and tracking.

We conducted two case studies using real-world video data. The software environment includes *JetPack 5.1.3-b29*, *Triton Inference Server r35.3.1*, *Redis Server 7.4.2*, and custom logic implemented in Python 3.11.

To optimize runtime efficiency, both models were converted into edge-optimized formats compatible with Triton. The trained YOLOv7-seg PyTorch model was converted into a TensorRT engine with FP16 precision, and the fine-tuned ConvNeXt model was exported as an ONNX model.

Two types of video sources were used: a fixed surveillance camera mounted on a transmission tower and aerial drone footage from a line patrol mission. The case studies demonstrate that the system can:
1) Accurately track fast-moving objects with appearance-aware identity matching.
2) Stably identify and track both rigid and non-rigid foreign objects.
3) Operate reliably on a low-cost edge device even at reduced frame rates.

*A.  Case I: Crane Vehicle Passing Through Transmission Corridor*

In this scenario, footage was captured by a fixed surveillance camera overlooking a transmission corridor. A central rectangular region beneath the transmission lines is defined as a critical zone—any foreign object entering this area is flagged as a potential threat. The experiment was conducted on April 11, 2023, in southern China, under cool spring-like weather conditions. Daytime temperatures ranged from 48 °F to 57 °F, with light rainfall and humidity levels between 70% and 75%. The lighting was generally diffuse due to overcast skies, and no artificial lighting was introduced. The crane vehicle moved at an approximate speed of 9–10 m/s. A fixed-position 1080p IP camera with a frame rate of 30 FPS and a horizontal field of view of 102° was used for video capture. It was installed at approximately 8–10 meters above ground, tilted downward at 35°.

As shown in Fig. 9, the YOLOv7-seg model successfully detected a crane vehicle entering the warning zone. The cropped object masks were passed to the ConvNeXt model to generate embeddings, and the vehicle's movement was tracked over six consecutive frames. The trajectory of its bounding boxes is visualized in Fig. 10.

Due to rapid motion, IoU scores between consecutive frames remained low. However, cosine similarity between the embeddings remained consistently high, ensuring robust identity tracking. Table 7 presents the IoU and cosine similarity scores across adjacent frames. Fig. 11 shows the feature comparison algorithm matches all frames to reference crane vehicle embeddings, confirming the foreign object as a crane vehicle.

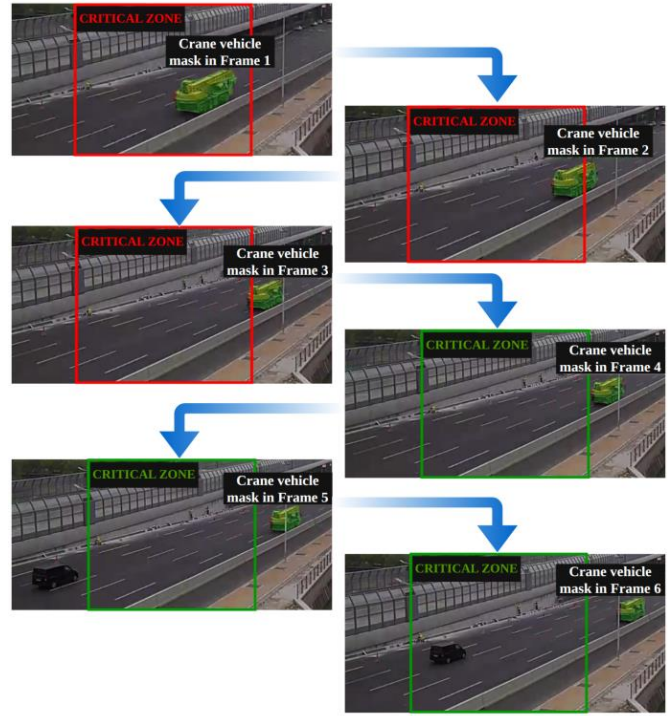

Fig. 9 Detection results showing a crane vehicle entering the critical warning zone.

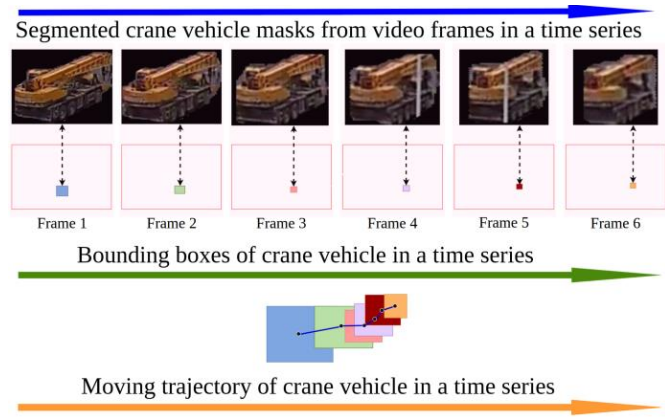

Fig. 10 Cropped crane vehicle masks and the trajectory across six frames.

TABLE 7 IoU AND COSINE SIMILARITY SCORES FOR CRANE

| Frame | 1-2 | 2-3 | 3-4 | 4-5 | 5-6 |
|---|---|---|---|---|---|
| IoU | 0.1568 | 0.3261 | 0.4375 | 0.3632 | 0.3420 |
| C-Score | 0.7789 | 0.7873 | 0.8470 | 0.7057 | 0.8829 |

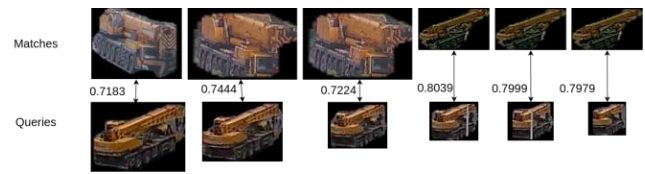

Fig. 11 Matched reference images and similarity scores for the crane vehicle across frames.

## B. Case II: Dust-Proof Net Entering Drone Clearance Zone

This scenario uses drone footage captured during a line patrol over a construction site. A rectangular region in the center of the image was defined as a flight clearance zone, and any object entering this area was considered a potential threat to drone safety. The experiment was conducted on a hazy day with visibility limited to approximately 4–5 km, further contributing to soft, diffused light. The drone used was a DJI Mavic Air 2, recording at 4K 25 FPS. Drone views provided top-down and oblique angles from 80–140 meters altitude.

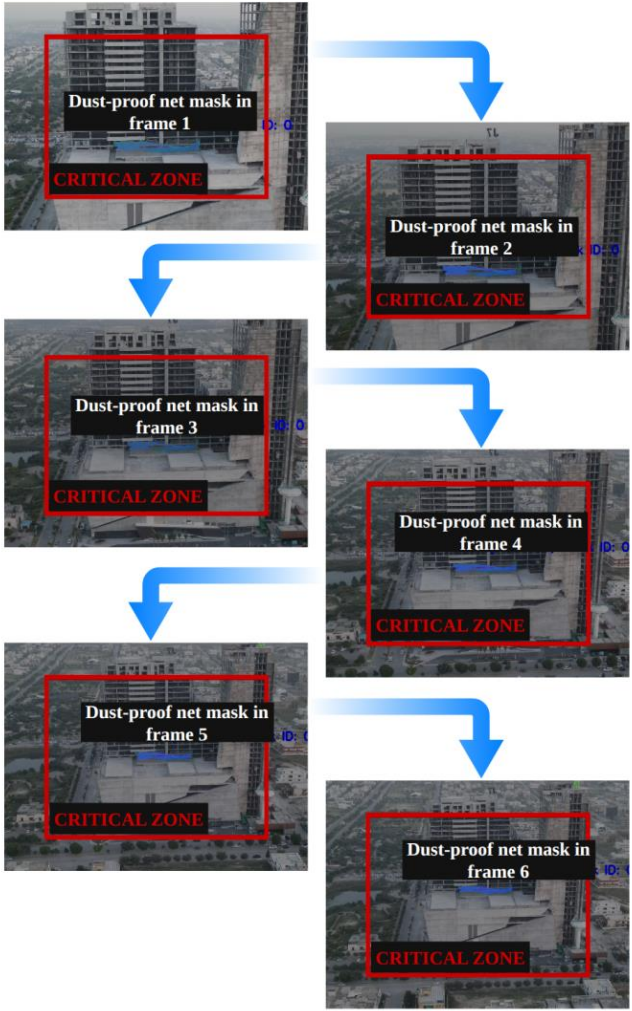

Fig. 12 Detection results from drone footage showing dust-proof net intrusion.

As shown in Fig. 12, both a tower crane and a dust-proof net were detected. Only the dust-proof net entered the restricted zone, triggering an FOI alert. Its bounding box trajectory across six frames is shown in Fig. 13, along with cropped object masks.

Due to the drone's movement and changing viewpoints, IoU scores were particularly low or even zero in several frames. Nevertheless, high cosine similarity scores between feature embeddings enabled successful object reassociation. The following table summarizes tracking performance:

All frames were correctly identified as dust-proof net instances as shown in Fig. 14, validating the system's ability to track non-rigid and deformable objects through the proposed feature-based algorithm despite motion-induced IoU degradation as shown in Table 8.

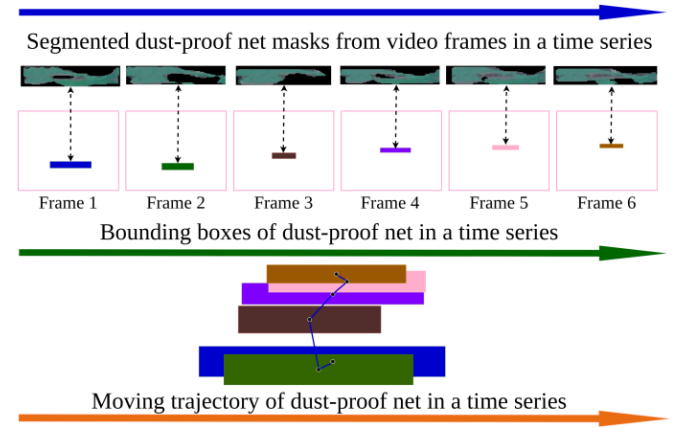

Fig. 13 Bounding box trajectory and cropped masks for the dust-proof net.

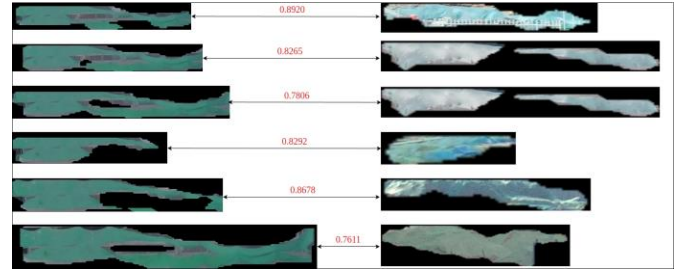

Fig. 14 Matched reference images for the dust-proof net across six frames.

TABLE 8 IoU AND COSINE SIMILARITY SCORES FOR DUST-PROOF NET TRACKING

| Frame | 1-2 | 2-3 | 3-4 | 4-5 | 5-6 |
|---|---|---|---|---|---|
| IoU | 0.4665 | 0 | 0 | 0.2375 | 0.4076 |
| C-Score | 0.7463 | 0.8008 | 0.9075 | 0.8300 | 0.8239 |

## C. Boundary Cases and Potential Failures

While the proposed system demonstrates strong overall performance in foreign object detection and tracking across diverse operational scenarios, several edge cases were observed during field testing that highlight current limitations. Notably, false alarms occasionally occurred when large clusters of deformable objects (e.g., greenhouse film or dust-proof net) exhibited uncertain detection results over consecutive frames, leading the model to misclassify them as moving intruding foreign objects. In addition, tracking inconsistencies were observed in situations involving multiple overlapping targets moving rapidly, where brief occlusions resulted in temporary identity switches. Furthermore, missed detections were more likely under low-visibility conditions, including overexposed frames or environmental glare, where the feature cosine similarity to reference embeddings dropped significantly—especially for deformable objects. These cases, while relatively infrequent, provide valuable insights into the boundary conditions of the system and indicate areas for future enhancement such as data and feature variety improvement, temporal consistency enforcement, and adaptive exposure handling.

V. CONCLUSION

This paper introduced a novel three-stage framework for real-time FOI detection and tracking in power transmission systems. By integrating YOLOv7 segmentation, ConvNeXt-based embedding with triplet loss, and a feature-assisted IoU tracker, the system achieves accurate detection and resilient tracking of both rigid and deformable objects under challenging conditions.

The contributions of this paper are threefold: (1) a decoupled design that enables retraining-free scalability to new FOI types, (2) robust performance validated on real-world surveillance and drone data, and (3) efficient, real-time deployment on affordable edge devices. These advances address long-standing challenges of unpredictability, deformability, and operational constraints in FOI monitoring.

Future work will focus on extending the system with fire detection, PTZ camera control for active monitoring, and large-scale field evaluations with utility partners, moving toward fully autonomous FOI monitoring to enhance power infrastructure safety and reliability.

VI. ACKNOWLEDGEMENT

This work was supported by the National Science Foundation (NSF) under award numbers 2418359 and 2301938.